\documentclass[final,5p,times,twocolumn]{elsarticle}

\usepackage{amssymb}
\usepackage{latexsym}

\usepackage{lineno,hyperref}
\usepackage{caption}
\usepackage{indentfirst}
\usepackage{graphicx}
\usepackage{float}
\usepackage{flushend}
\usepackage{stfloats}
\usepackage{booktabs} 
\usepackage{diagbox} 
\usepackage{multirow} 
\usepackage{tabu}
\usepackage{amsmath}
\usepackage{array}
\usepackage{soul}
\usepackage{color,xcolor}
\newcommand{\etal}{{\emph{et al.}}}
\usepackage{gensymb}
\usepackage{multirow}
\usepackage{threeparttable}
\usepackage{bbding}
\usepackage{pifont}
\usepackage{natbib}
\setcitestyle{numbers,square}

\biboptions{authoryear}
\journal{Computers in Biology and Medicine}

\begin{document}

\begin{frontmatter}



\title{A Rotation Meanout Network with Invariance for Dermoscopy Image Classification and Retrieval}

\author[mymainaddress]{Yilan Zhang}
\author[mymainaddress]{Fengying Xie\corref{mycorrespondingauthor}}
\cortext[mycorrespondingauthor]{Corresponding author}
\ead{xfy\_73@buaa.edu.cn}
\author[mysecondaryaddress]{Xuedong Song}
\author[mymainaddress]{Hangning Zhou}
\author[mymainaddress]{Yiguang Yang}
\author[mymainaddress]{Haopeng Zhang}
\author[mythirdaddress]{Jie Liu}

\address[mymainaddress]{Image Processing Center, School of Astronautics, Beihang University, Beijing 100191, China}
\address[mysecondaryaddress]{Shanghai Aerospace Control Technology Institute, Shanghai 201109, China}
\address[mythirdaddress]{Department of Dermatology, Peking Union Medical College Hospital, Chinese Academy of Medical Sciences and Peking Union Medical College, Beijing 100730, China}

\begin{abstract}
The computer-aided diagnosis (CAD) system can provide a reference basis for the clinical diagnosis of skin diseases. Convolutional neural networks (CNNs) can not only extract visual elements such as colors and shapes but also semantic features. As such they have made great improvements in many tasks of dermoscopy images. The imaging of dermoscopy has no principal orientation, indicating that there are a large number of skin lesion rotations in the datasets. However, CNNs lack rotation invariance, which is bound to affect the robustness of CNNs against rotations. To tackle this issue, we propose a rotation meanout (RM) network to extract rotation-invariant features from dermoscopy images. In RM, each set of rotated feature maps corresponds to a set of outputs of the weight-sharing convolutions and they are fused using meanout strategy to obtain the final feature maps. Through theoretical derivation, the proposed RM network is rotation-equivariant and can extract rotation-invariant features when followed by the global average pooling (GAP) operation. The extracted rotation-invariant features can better represent the original data in classification and retrieval tasks for dermoscopy images. The RM is a general operation, which does not change the network structure or increase any parameter, and can be flexibly embedded in any part of CNNs. Extensive experiments are conducted on a dermoscopy image dataset. The results show our method outperforms other anti-rotation methods and achieves great improvements in dermoscopy image classification and retrieval tasks, indicating the potential of rotation invariance in the field of dermoscopy images.
\end{abstract}


\end{frontmatter}


\section{Introduction}
Skin is the organ with the largest area and the closest contact with the outside world in the human body.  As a strong physiological barrier, the skin plays a key role in protecting human health \cite{yang2021microneedle}. Today, skin disease has become a common disease worldwide and a major public health problem \cite{siegel2019cancer}. Dermoscopy is a non-invasive microscopic image analysis technique that can magnify the area of skin lesions dozens of times, helping doctors to observe the fine structures and pigments under the skin surface \cite{vestergaard2008dermoscopy}. Thus, the use of dermoscopy images can significantly improve the diagnostic accuracy \cite{kittler2002diagnostic}. However, the diagnostic results are highly dependent on the experience of dermatologists and are less reproducible \cite{zhou2017multi}. Therefore, the demand for remote automatic diagnosis solutions has become more and more important with the increasing number of skin disease cases worldwide \cite{cassidy2022analysis}.

With the maturity of computer technology, the biomedical scientific community is increasingly interested in computer-supported skin lesion examination and characterization \cite{maglogiannis2009overview}. The computer-aided diagnosis (CAD) system can extract the features for tasks such as skin lesions classification and retrieval, to provide a more reliable diagnosis basis and have a far-reaching impact on dermatology clinic \cite{schmid2003towards}. Features with better separability can effectively represent dermoscopy data and eliminate irrelevant information, which largely determines the success of subsequent tasks. Hand-crafted features are early used\cite{mishra2016overview,xie2016melanoma}, and they are tipically low-level and lacked semantic information. During the past decade, the field of deep learning has developed vigorously. Among the studies, convolutional neural networks (CNNs) provide a learning-based solution for image feature extraction, which can effectively avoid the shortcomings of manual features and has been widely used in dermoscopy image analysis \cite{xie2020mutual,kawahara2016deep, yu2016automated, yao2021single, zhang2019attention,zhang2021dermoscopic}.

CNNs have significantly improved the performance in the tasks of dermoscopy image classification \cite{yao2021single,zhang2019attention}, retrieval \cite{allegretti2021supporting,zhang2021dermoscopic}, segmentation \cite{bi2017automatic} and so on with its ability of automatic feature learning. However, a challenging problem remains in dermoscopy images analysis: the orientation of skin lesion are diverse because the images are captured vertically from above, while the regular CNNs are not equivariant or invariant to the rotation\cite{Zhang_2017_ICCV}. That is, feeding rotated dermoscopy images to CNNs is not the same as rotating feature maps of the original image or not equivalently categorizing the images and their rotation versions \cite{han2021redet}. Thus the rotation of skin lesions in dermoscopy images would bring about adverse impacts on the feature learning process. For classification or retrieval tasks, the type of skin disease or the embeddings of features should not change due to the different dermoscopy imaging angles, which requires the CNNs to have certain invariance. In other words, the inference of the labels or the embeddings is not sensitive to image rotations.

Recent proposed methods extend CNNs to larger groups \cite{cohen2016group,zhou2017oriented,cheng2018rotdcf,zareapoor2021rotation,he2021efficient} which have additional orientation channels for different rotation angles. These methods usually replace the original layers of the network with newly designed convolutions, which introduce extra parameters and limit the modal generality for pre-training schemes \cite{kang2021rotation}. Besides, their applicabilities for dermoscopy image analysis remain to be verified. Therefore, this paper adopts the deep learning method to study the rotation-invariant feature extraction of dermoscopy images. We proposed a novel general operation, namely rotation meanout (RM), which does not change the network structure or increase the parameters. The proposed method expands the input feature maps by rotating them at equal interval angle, and then fuses these expanded feature maps using meanout operation.  Extensive experiments conducted on a  representative images dataset provided by International Skin Imaging Collaboration (ISIC) 2019 classification challenge \cite{tschandl2018ham10000,codella2018skin,combalia2019bcn20000}, called ISIC 2019 dataset.  We analyzed the factors influencing the performance of RM and verified that rotation invariance features can be effectively extracted, which can help get superior performance in classification and retrieval tasks. 

In summary, the contributions of this research are three-fold: 
\begin{itemize}
\item We propose a new network for dermoscopy image feature extraction, RM, to solve the problem of a large number of skin lesion target rotations in the dermoscopy dataset. It is theoretically proven that RM is rotation-equivariant, even rotation-invariant when being followed by the global average pooling (GAP) operation. 
\item The proposed RM network does not change the network structure or introduce extra parameters. It can be flexibly combined with any part of a CNN and effectively improves the ability of rotation invariance. 
\item We verified the effectiveness of the RM in dermoscopy image classification and retrieval tasks, respectively, demonstrating its validity and generality. Experiments on the dermoscopy image dataset show that our method can achieve superior performance than other rotation invariance or equivariance methods commonly used for natural images. Compared with other skin lesion classification and retrieval methods, RM also gets satisfactory performance.
\end{itemize}
The rest of this paper is organized as follows: Section 2 lists the related works. Section 3 gives the detailed description of the proposed RM network and the theoretical analysis of rotation invariance. Section 4 lists the experiments to verify the effectiveness of the RM operation. And the conclusion is presented in section 5.

\section{Related Works}

\subsection{Dermoscopy Image Feature Extraction Based on CNNs}
CNNs with hierarchical feature learning capability have made significant advances in dermoscopy image feature extraction in recent years. Kawahara \etal \cite{kawahara2016deep} presented a fully CNN based on AlexNet \cite{krizhevsky2012imagenet} to extract representative features of melanoma. Yu \etal \cite{yu2016automated} used ResNet34 \cite{he2016deep} as the network architecture in the classification stage. With the continuous proposal of new network structures, SENet \cite{hu2018squeeze}, EfficientNet \cite{tan2019efficientnet}, RegNet\cite{radosavovic2020designing}, etc. can also be used as feature extraction network \cite{yao2021single}. Although these methods have achieved better performance than those based on hand-crafted features, they are mainly designed for natural images such as ImageNet. Some studies have improved the network structure for dermoscopy image characteristics, thereby enhancing the discriminative representation ability of CNNs. Zhang \etal \cite{zhang2019attention} proposed an attention residual learning (ARL) mechanism to make the network focus on the area of skin lesions. Another proposed attention model called hybrid dilated convolution spatial attention (HDCSA) is inserted into the network, which can not only extract global features but also detailed features of skin lesions \cite{zhang2021dermoscopic}. However, the methods above do not account for the fact that skin lesions have diverse orientations in dermoscopy images.

\subsection{Rotation-invariant Features of CNNs}
Good features should be invariant in image classification and retrieval tasks \cite{sohn2012learning}. This is because in the classification task, the category label of the image does not change as the position of the classified object changes, and in the retrieval task, the distance between features of the intra-class is expected to be smaller, even if there are some image transformations.  Local connection, weight sharing, and pooling operation in CNNs can provide translation invariance\cite{kanazawa2014locally} and the multilayer structure of CNNs expands the receptive field to provide scale invariance. However, CNNs are not invariant to the rotations, resulting in a weak generality. To overcome this limitation, many works on invariant features based on CNNs have been carried out. The majority of them can be divided into three main strategies: (i) data augmentation, (ii) rotated filters, and (iii) anti-transformation.

Data augmentation (DA) \cite{van2001art} is a typical approach to firstly improve the rotation invariance of the CNNs. DA-based methods can learn rotation invariance features by simply rotating input images at random orientation, among which, TI-Pooling \cite{laptev2016ti} use the multiple rotated images as input and fuse the outputs using a pooling strategy, where a parallel siamese network is employed to extract the features. While some works also utilize siamese CNNs and learn the representation from a loss function perspective\cite{kang2021rotation,qi2021rotation}. Another idea is to exploit polar mapping to convert rotation to translation of the input images \cite{kim2020cycnn}. However, DA-based methods focus on learning the features from the enriched inputs with the risk of overfitting as well as high computational complexity\cite{weiler2018learning}. The second kind of method usually uses a bank of rotated filters to obtain the feature maps of different orientation channels to obtain the rotation-equivariant or rotation-invariant features. For example, Marcos \etal \cite{marcos2016learning} encode rotation invariance by simply rotating the convolutional kernels. Cohen \etal \cite{cohen2016group} introduce group convolutions (GCNNs), which repeat the transformed filters at different rotation-flip combinations. Zhou \etal \cite{zhou2017oriented} propose an oriented response network (ORN) that used active rotating filters (ARFs) to explicitly encode the location and orientation information of feature maps. After that, some new methods were raised on this basis \cite{cheng2018rotdcf,zareapoor2021rotation,he2021efficient}. The methods based on rotated filters usually use new convolution layers to replace the traditional ones. However, the cost is that they introduce extra parameters and limit the generality of models due to the need for pre-training schemes. On the other hand, the sizes of filters are usually small and features are easily influenced by interpolation errors. Anti-transformation methods\cite{jaderberg2015spatial,dai2017deformable} take rotation as a special case of transformation. For example, spatial transformation networks (STN) \cite{jaderberg2015spatial} and deformable convolutional networks (DCN) \cite{dai2017deformable} introduce a new module to learn transformation without any extra training supervision. Though they are widely used to get rotation invariant features \cite{han2021redet}, the problem of the complex transformation of parameters is not solved well to data. Besides, they are not especially for rotation and do not have strictly invariant and also need extra parameters.

To tackle the issues above, we propose a novel rotation meanout (RM) network with invariance for dermoscopy image classification and retrieval tasks. Compared with the DA-based method, our method does not need to rotate the images, which can be more computationally efficient. While compared with the rotated filters and anti-transformation methods, it does not increase any parameters of the network, and the RM is more general and can be applied to any part of a classic CNN. Table \ref{saotas} summarizes the properties of the state-of-the-arts mentioned above. 

\begin{table*}[!t]
\centering
\caption{\label{saotas}State-of-the-arts in the domain of rotation-invariant feature: summary of main properties}
\renewcommand{\arraystretch}{1.1}
\begin{tabular}{lcccl}
\hline
Approach                 & \begin{tabular}[c]{@{}c@{}}Change Network\\ Structure\end{tabular} & \begin{tabular}[c]{@{}c@{}}Introduce Extra\\ Parameters\end{tabular} &  Strategy                                 & \multicolumn{1}{l}{Characteristic \& Limitation}                                                                                                           \\ \hline
TI-pooling \cite{laptev2016ti}               & Yes                                                                 & No                                                                                                                             & \multirow{3}{*}{Data Augmentation}       & \multirow{3}{*}{\begin{tabular}[c]{@{}l@{}}Rotate input images;\\ Expensive computational.\end{tabular}}                                               \\
RiDe  \cite{kang2021rotation}                   & No                                                                 & No                                                                                                                                &                                          &                                                                                                                                                            \\
RIR  \cite{qi2021rotation}                    & No                                                                 & No                                                                                                                                 &                                          &                                                                                                                                                            \\ \hline
GCNN  \cite{cohen2016group}                   & Yes                                                                & Yes                                                                                                                              & \multirow{3}{*}{Rotated Filters}         & \multirow{3}{*}{\begin{tabular}[c]{@{}l@{}}A bank of oriented filters;\\Introduce interpolation errors;\\ Low generality.\end{tabular}}        \\
ORN \cite{zhou2017oriented}                     & Yes                                                                & Yes                                                                                                                             &                                          &                                                                                                                                                            \\
E$^4$-Net \cite{he2021efficient} & Yes                                                                & Yes                                                                                                                               &                                          &                                                                                                                                                            \\ \hline
STN  \cite{jaderberg2015spatial}                    & Yes                                                                & Yes                                                                                                                                & \multirow{2}{*}{Anti-transformation}     & \multirow{2}{*}{\begin{tabular}[c]{@{}l@{}}Take rotation as a special case;\\ Not strictly rotation-invariant.\end{tabular}} \\
DCN  \cite{dai2017deformable}                   & Yes                                                                & Yes                                                                                                                              &                                          &                                                                                                                                                            \\ \hline
RM(Ours)                 & No                                                                 & No                                                                                                                               & \multirow{1}{*}{Rotated Feature Maps} & \begin{tabular}[c]{@{}l@{}}Rotation small-size feature maps;\\ High generality.\end{tabular}                                                                \\ \hline
\end{tabular}
\end{table*}

\subsection{Rotation-invariant Features for Dermoscopy Images}
In recent years, a large number of research papers based on rotation equivariance and variance have made success in the fields including remote sensing images \cite{qi2021rotation}, 3D point clouds\cite{xu2021sgmnet}, face detection \cite{zhou2022mtcnet}, molecular representation \cite{li20213dmol}, etc. However, few scholars have gradually noticed the importance of rotation-invariant features in the field of dermoscopy image tasks. Li \etal \cite{li2018deeply} introduced GCNNs \cite{cohen2016group}, and proposed G-upsampling and G-downsampling for the task of skin lesion segmentation. This method makes the network have a certain rotation equivariance but may introduce interpolation errors and extra parameters. Based on the rotation-invariant deep hashing network \cite{zhang2021dermoscopic}, the Cauchy rotation loss function is designed, which is mainly aimed at the retrieval task of dermoscopy images, and the method of rotating the input has an expensive computational cost. Despite the success of rotation invariance studies in natural image tasks, research on rotation-invariant features in dermoscopy images still has great potential.

\section{Methodology}
CNNs lack rotation invariance \cite{laptev2016ti}. Although many scholars have made some progress in this field, their methods have problems such as increasing the number of model parameters and changing the network structure, and the applicability of the dermoscopy image task remains to be verified. To overcome these limitations, we proposed a new network called rotation meanout (RM), to improve the feature extraction ability on rotations of the networks. Our proposed RM does not change the network structure or increase parameters, which is easily implemented. Besides, RM is a general model and can be flexibly combined with any part of a CNN.
\subsection{The Proposed RM Network}
Let $ X=[x_{0},x_{1},...,x_{n-1}]$ be a set of input feature maps with numbers of $n$, the proposed RM model can be described as follows:
\begin{equation}
\begin{aligned}\label{e1}
&RM(X)=p[f(X), r_{\theta}^{-1}f(r_{\theta}X), r_{2\theta}^{-1}f(r_{2\theta}X), \dots, r_{(k-1)\theta}^{-1}f(r_{(k-1)\theta}X)]  
\end{aligned}
\end{equation}
Where $ f(\cdot) $ represents a convolution operation to extract features, $ r_{i\theta}(i\in[0, k)) $ represents rotating $ i\cdot\theta $ degrees clockwise and $ r_{i\theta}^{-1} $ represents rotating $ i\cdot\theta $ degrees counterclockwise, $ k $ denotes the number of rotation and is subject to $ k\cdot\theta=360^\circ $, $ p $ stands for a fusion operation.

\begin{figure*}[!t]
	\centering
	\includegraphics[width=1\textwidth]{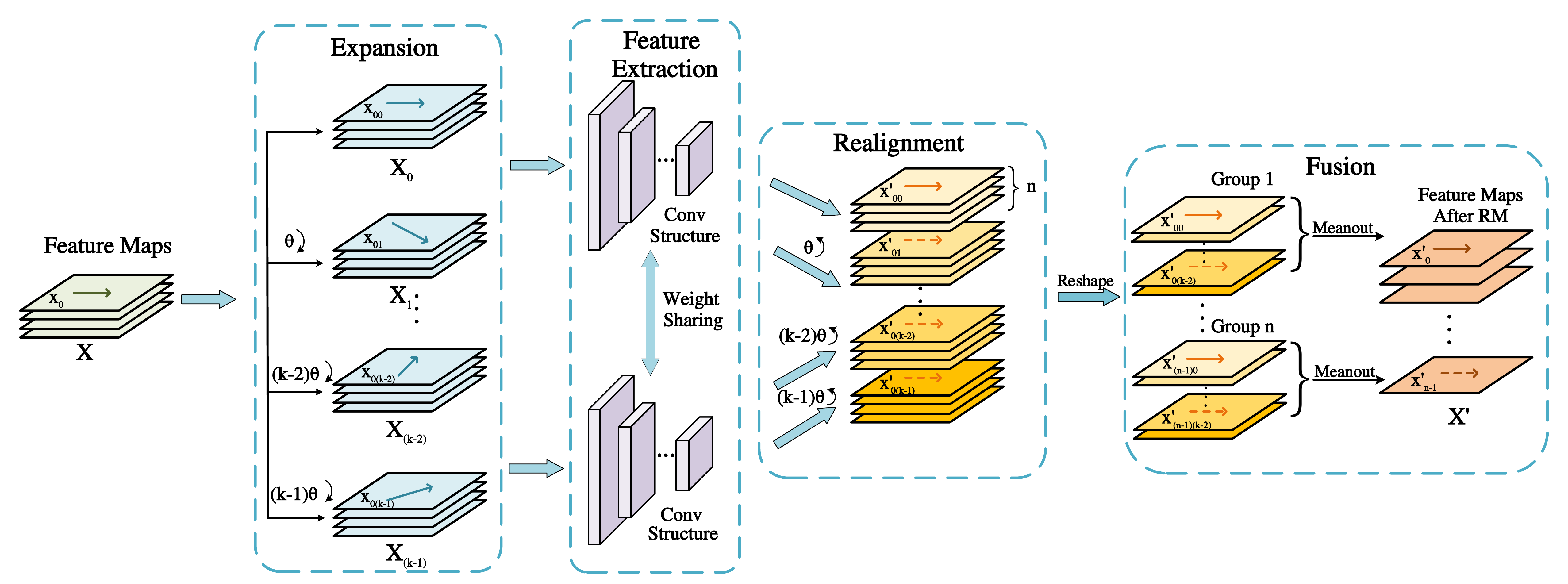}
	\caption{RM network. In expansion stage, feature maps are expanded $k$ times by rotating them at same interval angles $\theta$. In realignment stage, the $k$ groups of rotated feature maps are input to a shared convolutional structure to extract features. Then, inversely rotate the feature map back to its original angle, so that the outputs can be realigned to keep their spatial position relation. Finally, these $k$ sets of realigned feature maps are fused to obtain one set of final feature maps by using the mean pooling operation.}
	\label{f1}
\end{figure*}

Figure \ref{f1} shows the flow chart of the RM network, which includes four steps: expansion, feature extraction, realignment, and fusion.
Firstly, feature maps are expanded $ k $ times by rotating them at equal interval angles. Different from data rotation augmentation which rotates the original images, here we rotate the feature maps by $ i\cdot\theta (i = 0, 1, ..., k-1) $ degrees before feeding them into the convolution layers. Interpolation is needed to complete the rotation operation for the feature maps. When the interpolation coordinate is not an integer, an interpolation error will be introduced. Usually, as the depth of the CNNs increase, the size of the feature maps will continuously decrease, and the error will be more obvious. 

Secondly, the obtained $ k $ sets of rotated feature maps are sequentially input to a shared convolutional structure to extract features. This convolutional structure can be a simple or several convolutional layers, and even any more complex structure like an Inception Block in GoogLeNet \cite{szegedy2016rethinking} and a residual block in ResNet\cite{he2016deep}, etc. At the same time, We utilize the weight sharing strategy during the process of training the shared convolutional structure.

Thirdly, the $ k $ sets of new feature maps output are realigned to keep their spatial position relation. Formula \ref{e1} can be written as the multiplied form of two vectors: 
\begin{equation}\label{e2}
\begin{split}
&RM(X)=p[[1, r_{\theta}^{-1}, r_{2\theta}^{-1}, \dots, r_{(k-1)\theta}^{-1}]\circ
[f(X), f(r_{\theta}X), f(r_{2\theta}X),\\ &\dots,f(r_{(k-1)\theta}X)]]
\end{split}
\end{equation}
In (\ref{e2}), vector $ [1, r_{\theta}^{-1}, r_{2\theta}^{-1}, \dots, r_{(k-1)\theta}^{-1}] $ is exactly the realign operation. That is, after the convolutions, the $ i^{th}  (i\in[0, k)) $ set of rotated feature maps are realigned through rotating correspondingly counterclockwise by $ i\cdot\theta $.

Finally, we use the orientation mean pooling operation to fuse these $k$ sets of realigned feature maps into a single set of final feature maps:
\begin{equation}\label{e3}
p_{mean}(X)=mean[X_0^\prime, X_1^\prime, \dots, X_{k-1}^\prime]
\end{equation}
where $ X_i^\prime (i\in[0,k)) $ represents the $ i^{th} $ set of feature maps after realignment step. This set of final feature maps has the same number of channels as the shared convolutional structure's original input, which allows the network structure to remain constant when the RM is embedded in a baseline network. 
\subsection{Analysis of RM}
\subsubsection{Rotation Equivariance and Rotation Invariance}
It can be proved that the proposed network is rotation-equivariant to the input feature maps $ X $. Rotation equivariance means that when the input is rotated, the learnt feature maps also change in a predictable way \cite{lenc2015understanding}. More formally, a function $ f $ is equivariant to a class of transformations $T$. For transformation $ t \in T $ of the input $ X $, if a corresponding transformation $ t^\prime $ of the output $ f(X) $ can be found, $ f(tX)=t^{\prime}f(X) $ for all $ X $ \cite{schmidt2012learning}. From (\ref{e2}) we have:

\begin{equation}\label{e4}
\begin{split}
&RM(r_{\theta}X)\\
& =p[f(r_{\theta}X), r_{\theta}^{-1}f(r_{2\theta}X), r_{2\theta}^{-1}f(r_{3\theta}X), \dots, r_{(k-1)\theta}^{-1}f(r_{k\theta}X)]\\
& =r_{\theta}\cdot p[r_{\theta}^{-1}f(r_{\theta}X), r_{2\theta}^{-1}f(r_{2\theta}X), r_{3\theta}^{-1}f(r_{3\theta}X), \dots,
r_{k\theta}^{-1}f(r_{k\theta}X)
\end{split}
\end{equation}
$ k\cdot\theta=360^\circ $, therefore $ r_{k\theta}^{-1}f(r_{k\theta}x)=f(X) $, and (\ref{e4}) can be rewrite as: 
\begin{equation*}
RM(r_{\theta}X)=r_{\theta}\cdot RM(X)
\end{equation*}
Thus we have:
\begin{equation}\label{e5}
RM(r_{i\theta}X)=r_{i\theta}\cdot RM(X),i\in[0,k)
\end{equation}
From (\ref{e5}), the proposed RM is rotation-equivariant.

Since many CNNs have a global average pooling (GAP) operation before the fully connected layers, which enforces each feature map to be a feature point and discards all of the spatial information. Let $ GAP(\cdot) $ denote the GAP operation, we have:
\begin{equation}\label{e6}
GAP(RM(r_{i\theta}X))=GAP(r_{i\theta}\cdot RM(X))=GAP(RM(X))
\end{equation}

Therefore, the GAP operation in classification and retrieval tasks makes the proposed RM network have rotation invariance, which improves the robust of influence by the angle changes of input images. 

\subsubsection{Weight Sharing Strategy}
In CNNs, the weight sharing of a convolution kernel is used to calculate convolution on an original image or on a set of feature maps, which reduces the parameters, and more importantly, the mined features have the ability of anti-translation. In the proposed RM network, all sets of the rotated feature maps are sequentially input to the convolutional structure and the weights are shared. The weight sharing strategy does not change the original convolutional structure or increase any parameter. Furthermore, because of the strategy, RM has the property of rotation equivariance and rotation invariance.

\subsection{Combine an RM with a CNN}
\begin{figure}[!t]
	\centering
	\includegraphics[width=0.48\textwidth]{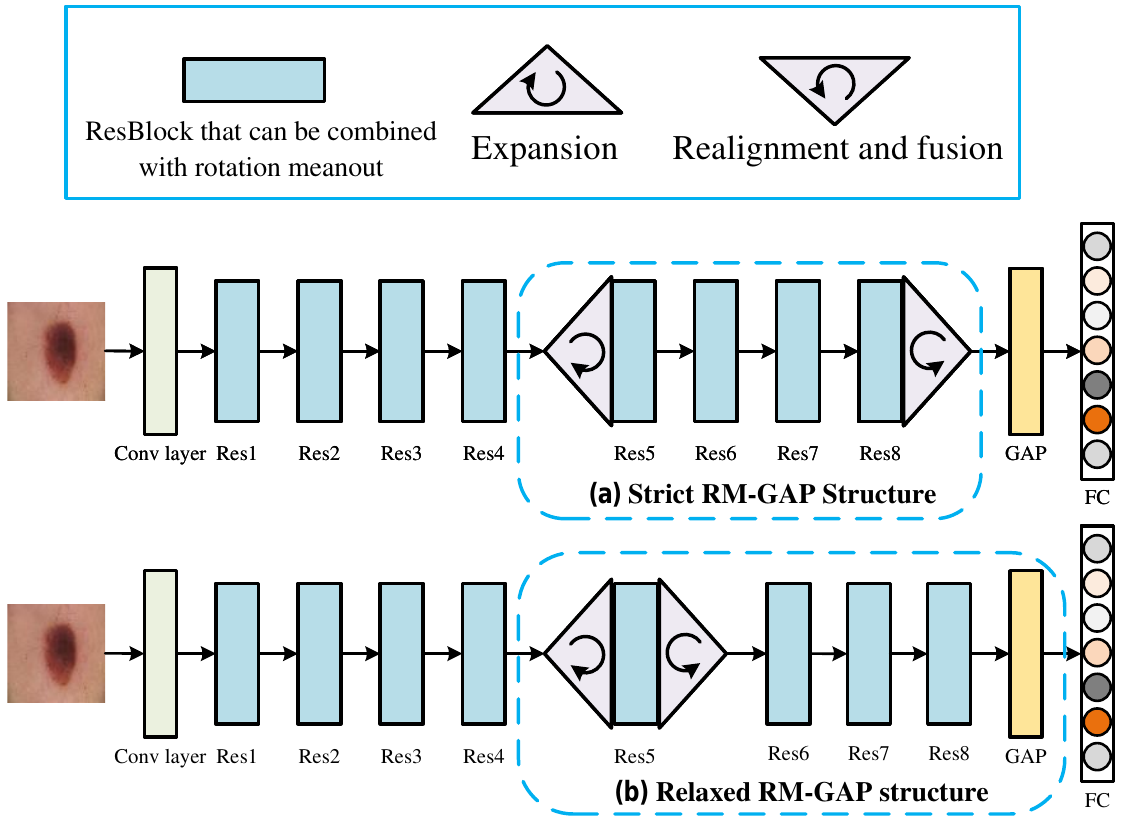}
	\caption{Embedding an RM in ResNet18.}
	\label{f2}
\end{figure}

The proposed RM does not change the dimensions of the input feature maps and can be easily embedded in a CNN. Figure \ref{f2} shows two examples of embedding an RM operation in ResNet18 \cite{he2016deep} (ResNet18 contains 8 residual blocks). In Figure \ref{f2} (a), four residual blocks from Res 5 to Res 8 are taken as the shared convolutional structure to be combined with the RM network. It is a strict RM-GAP structure (the RM is directly followed by the GAP operation), and the extracted features are rotation-invariant to the input feature maps of Res 5. In Figure \ref{f2} (b), only Res 5 is combined with the RM. It is a relaxed RM-GAP structure (the RM is separated from the GAP operation).
Due to the three residual blocks between Res 5 and GAP are not rotation-equivariant, the relaxed RM-GAP structure is not rotation-invariant. However, it still with the good anti-rotation ability and can improve the network performance, which will be verified in our experiments. The proposed RM is a general model, and any convolutional block in CNNs can be taken as the basic unit and combined with an RM model. 

\section{Experiments}
\subsection{Experimental Setup}
\subsubsection{Dataset}
\begin{figure*}[!t]
	\centering
	\includegraphics[width=0.9\textwidth]{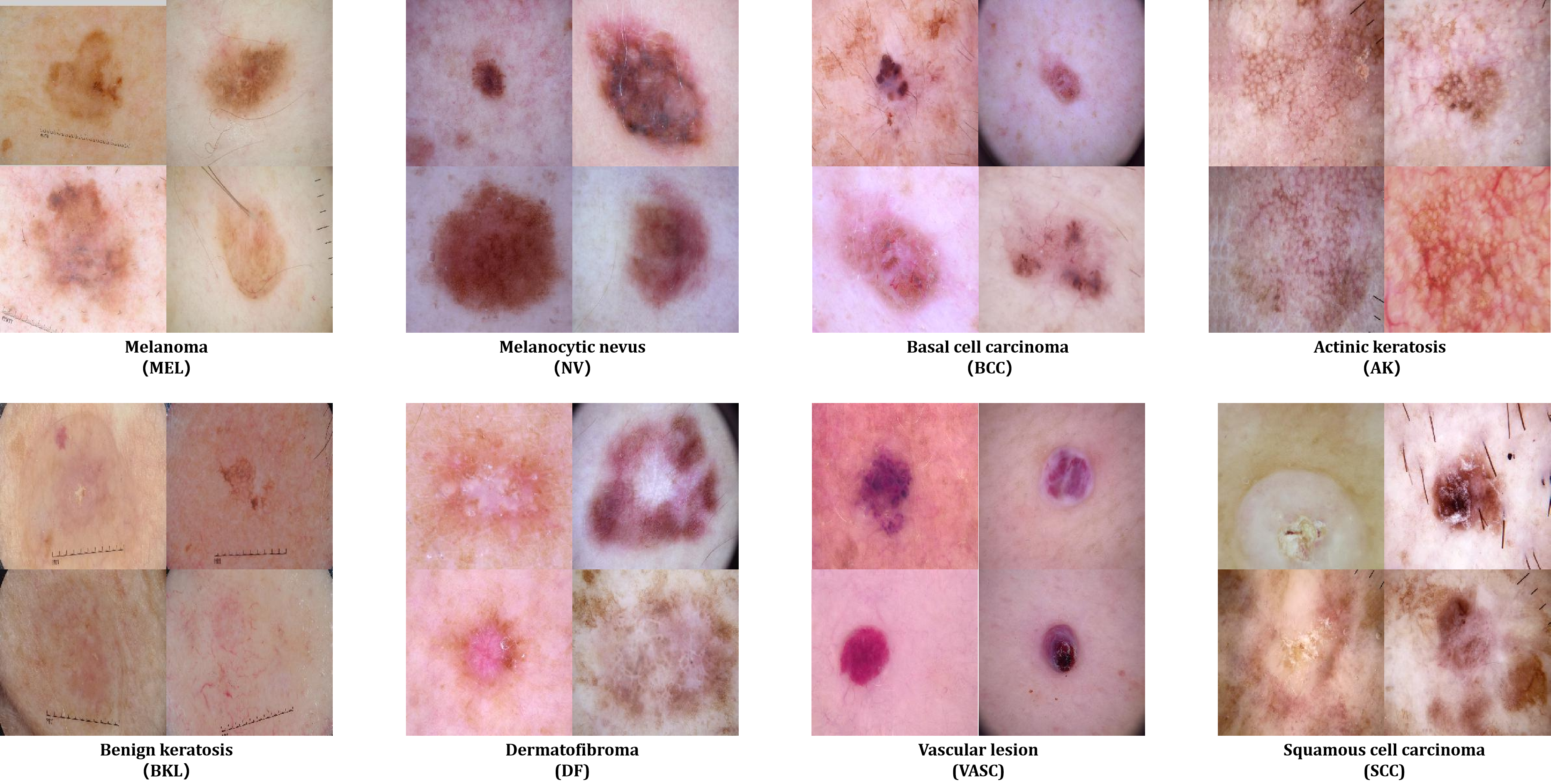}
	\caption{Examples of eight lesion classes in the ISIC2019 dataset.}
	\label{ISIC_fig}
\end{figure*}

We evaluate the performance of our proposed method on a dermoscopy images datasets provided by the International Skin Imaging Collaboration (ISIC) 2019 classification challenge (ISIC 2019). The open dermoscopy images dataset consists of 25,331 images across 8 different diagnostic categories with details in Table \ref{ISIC2019} and the examples are shown in Figure \ref{ISIC_fig}. We divide the dataset into the training set, validation set, and testing set according to the ratio of 8:1:1. In this study, each original image in the training set is scaled to 256$ \times $256 by short edge, and a 224$ \times $224 crop is randomly sampled from it or its horizontal flip or vertical flip, with the per-pixel RGB scale to (0, 1) and mean value subtracted and standard variance divided. It needs to be indicated that rotation augmentation is not used here, to avoid the correlation between rotation augmentation and our RM network.

\newcommand{\tabincell}[2]{\begin{tabular}{@{}#1@{}}#2\end{tabular}}
\begin{table}[!t]
	\centering
	\caption{\label{ISIC2019}Details of the ISIC 2019 dataset.}
	\renewcommand{\arraystretch}{1.1}
	\setlength\tabcolsep{2.0pt}
	\begin{tabular}{clcl}
		\hline
		Category & Description                                                                           & Sample Number & Proportion \\ \hline
		MEL      & Melanoma                                                                              & 4522          & 17.85\%    \\
		NV       & Melanocytic nevus                                                                     & 12875         & 50.83\%    \\
		BCC      & Basal cell carcinoma                                                                  & 3323          & 13.12\%    \\
		AK       & Actinic keratosis                                                                     & 867           & 3.42\%     \\
		BKL      & Benign keratosis  & 2624          & 10.36\%    \\
		DF       & Dermatofibroma                                                                        & 239           & 0.94\%     \\
		VASC     & Vascular lesion                                                                       & 253           & 1.00\%     \\
		SCC      & Squamous cell carcinoma                                                               & 628           & 2.48\%     \\ \hline
	\end{tabular}
\end{table}

\subsubsection{Implementation Details}
We consider four CNNs, including ResNet18, ResNet34, GoogleNet-Inception-V3\cite{szegedy2016rethinking} and EfficientNet-b0\cite{tan2019efficientnet} as baselines. To evaluate the performance improvement of the proposed RM network, we conducted experiments on both classification and retrieval tasks. Since hashing method is one of the most widely used methods for image retrieval, we refer \cite{lin2015deep} to adopt the deep hashing method. Thus, we modify the baselines by adding a fully-connected layer containing $L$ units for hash code learning before the classification head. The bit of hash code $L$ is set to 16 and the cosine distance is computed between the query image and each point in the database. For two tasks, we all use the weighted cross-entropy as the loss function. Besides, transfer learning is used to initialize the CNNs with the parameters pre-trained on the ImageNet dataset. 

The networks are trained on a PC with two Nvidia GeForce GTX 1080Ti GPUs. The whole work is implemented with the open-source library named PyTorch. We utilize the mini-batch stochastic gradient descent (SGD) optimizer to train the CNN models with an initial learning rate of 0.01 and a decay rate of 0.1 every 20 epochs. The batch size is set to 32, and the weight decay and momentum are set to 0.0005 and 0.9, respectively. We train the networks for a total of 60 epochs for classification task and 80 epochs for retrieval task. The results of the experiments are averaged multiple times and the standard deviations (SD) are reported. The codes of this paper can be publicly available from \url{https://github.com/vemvet/Rotation-Meanout-Network.git}.

\subsubsection{Evaluation Metrics}
\textbf{Classification Task.} We use precision, sensitivity, specificity, and Kappa coefficient to evaluate the performance of classification task, as shown in the following formula respectively:

\begin{equation}\label{Precision}
Precision = \dfrac{TP}{TP+FP}
\end{equation}

\begin{equation}\label{Sensitivity}
Sensitivity = TPR = \frac{TP}{TP+FN}
\end{equation}

\begin{equation}\label{Specificity}
Specificity = 1-FPR = 1-\dfrac{FP}{FP+TN} = \dfrac{TN}{FP+TN}
\end{equation}
where $TPR$ means true positive rate, $FPR$ means false positive rate. $TP$, $FN$, $TN$ and $FP$ represent the number of true positive, false negative, true negative, and false positive, respectively, and $M$ denotes the number of classes. The higher the precision, sensitivity, and specificity, the better the classification performance of the algorithms. The averaged value of each metric is reported for the evaluation, denoted as AP, Ave Sen, and Ave Spec.

 Moreover, we also use the Kappa coefficient \cite{hunt1986percent} to measure the consistency of classification accuracy between results of different algorithms and it is derived as:

\begin{equation}\label{Kappa}
Kappa = \frac{p_{0}-p_{e}}{1-p_{e}}
\end{equation}
where $p_{0}$ is the overall classification accuracy, $p_{e}$ is the hypothetical probability of a chance agreement.  Assuming that the numbers of real samples in each category are $a_{1}, a_{2},...,a_{C}$ and the predicted numbers of samples in each category are $b_{1}, b_{2},...,b_{C}$, respectively, and the total number of samples is $n$ , then $p_{0}$ and $p_{e}$ can be expressed as:

\begin{equation}\label{p0}
p_{0} = \dfrac{TN+TP}{TN+FP+FN+TP}
\end{equation}

\begin{equation}\label{pe}
p_{e}=\dfrac{a_{1}\times b_{1}+a_{2}\times b_{2}+...+a_{C}\times b_{C}}{n\times n}
\end{equation}

When Kappa is greater than 0.75, the model has good consistency, when Kappa is between 0.40 and 0.75, it represents moderate consistency, and when Kappa is less than 0.40, it indicates that the model has poor consistency.

\textbf{Retrieval Task.} We use the mean average precision (mAP) and mean reciprocal rank (mRR) to evaluate the performance, as shown in (\ref{mAP}) and (\ref{mRR}). The top 10 dermoscopy images in the retrieval stage are used as the final retrieval results.

\begin{equation}
\label{mAP}
\begin{split}
 mAP@10 & =\frac{1}{M}
\sum_{c=1}^{M}\frac{1}{|Q_{c}|}\sum_{i=1}^{Q_{c}}\frac{TP_{i}}{TP_{i}+FP_{i}}\\
& =\frac{1}{M}
\sum_{c=1}^{M}\frac{1}{|Q_{c}|}\sum_{i=1}^{Q_{c}}\frac{TP}{10}  
\end{split}
\end{equation}

\begin{equation}
\label{mRR}
mRR@10=\frac{1}{M}
	\sum_{c=1}^{M}\frac{1}{|Q_{c}|}\sum_{i=1}^{Q_{c}}\frac{1}{rank_{i}}
\end{equation}
where $Q_{c}$ is a set of query images with category $c$ in the test set and $|Q_{c}|$  is the number of elements in the set. $rank_{i}$ is the ranking of the first correct matching image when retrieving the $i-th$ image.

\subsection{Verification of the Proposed Structure}
In this part, We use ResNet18 as the baseline and the framework in Figure \ref{f2} (a) to verify the effectiveness of the proposed RM structure in skin lesion classification task.

\subsubsection{Ablation Study}
We probe the effect of the proposed RM by ablation experiments. Table \ref{Ablation} presents the results of the ablation study, where RM represents the strict RM-GAP structure in Figure \ref{f2} (a), RM-WR means no rotation operations in the expanding and realign stages, RM-NWS denotes without weight sharing in the RM (each of rotation branches has an independent convolutional structure). It can be seen that module without rotation operation and the baseline have a similar performance within training error. For RM-NWS, the extracted features do not have rotation invariance, so the classification effect is less improved. Compared with the baseline, the module containing the rotation operation and using the weight sharing strategy has a significantly improved classification performance of baseline, in which the AP is increased by 9.25\%, and the Ave Sen is increased by 3.90\%. In addition, weight sharing has less local parameters, and thus significantly reduces model size.

\begin{table}[!t]
	\centering
	\caption{\label{Ablation}Ablation study on RM}.
	\renewcommand{\arraystretch}{1.2}
	\begin{tabular}{lccc}
		\hline
		Network                & \multicolumn{1}{c}{AP (\%)} & \multicolumn{1}{c}{Ave Sen (\%)} & \multicolumn{1}{c}{Ave Spec (\%)} \\ \hline
		baseline               & 55.41$\pm$0.95                       & 72.07$\pm$1.57                                       & 96.78$\pm$0.17                                        \\
		RM-WR & 55.86$\pm$1.10                       & 73.88$\pm$2.70                                        & 96.83$\pm$0.10                                        \\
		RM-NWS    & 56.31$\pm$2.23                       & 74.19$\pm$1.08                                        & 96.99$\pm$0.08                                        \\
		RM   & \textbf{64.66$\pm$0.99}                       & \textbf{75.97$\pm$0.81}                                        & \textbf{97.16$\pm$0.04}                                       \\ \hline
	\end{tabular}
\end{table}

\begin{figure*}[!t]
	\centering
	\includegraphics[width=1\textwidth]{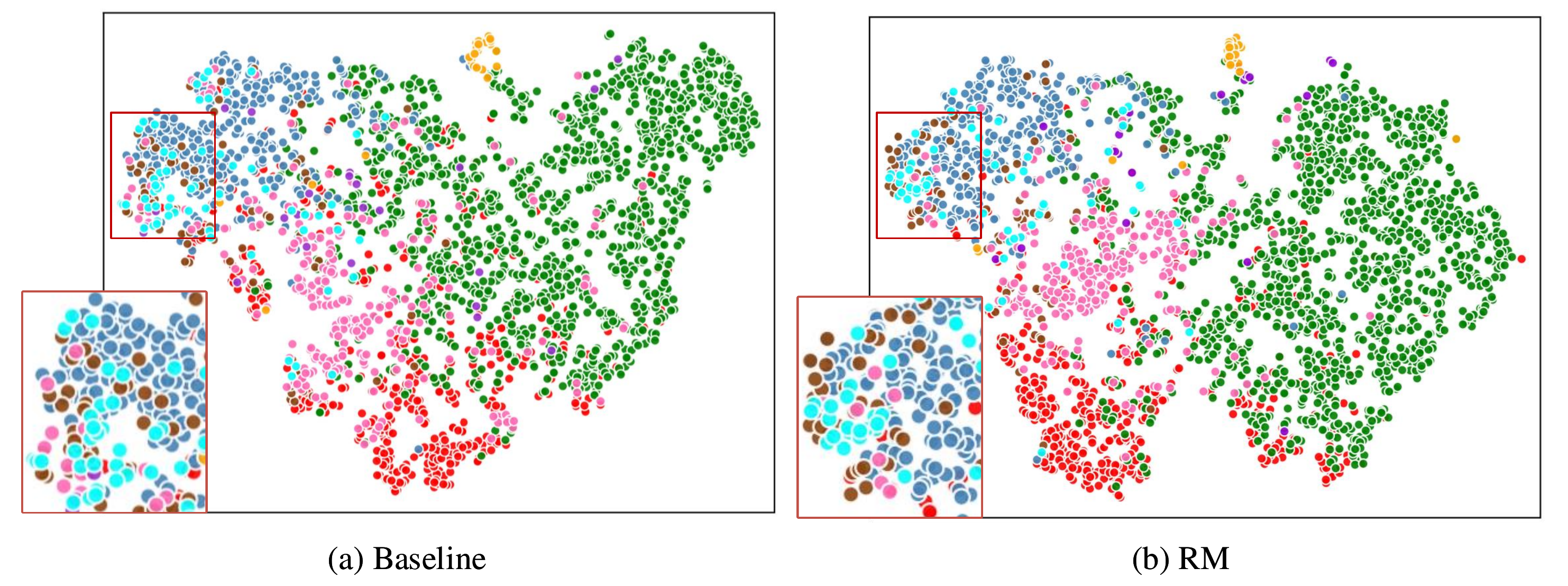}
	\caption{T-SNE visualization of the last layer representations of the networks, where (a) is the visualization of ResNet18 and (b) is that of ResNet18 combined with RM. (Refer to the legend in Figure \ref{f5} for the corresponding colors of skin diseases)}
	\label{f4}
\end{figure*}

We use the output of the classification layer as the features of the networks, and use the t-SNE\cite{van2008visualizing} method to visualize it on the testing set. As we can see from Figure \ref{f4}, with RM, the distribution of the features is more dispersed in different categories and more aggregated within the same category, and the pink part is particularly obvious. To keep loss low during training, the sample distribution of minority classes will be scattered and the classifier performs worse. However, it can be seen from the red box that the RM can compact the features of minority classes (color cyan). Table \ref{table8} shows the metrics of each category, the precision of SCC increased from 23.12\% of baseline to 40.86\%, and the sensitivity increased from 63.24\% to 64.72\%. In summary, RM can improve the separability of features to a certain extent, whether the classes of majority or minority.

\subsubsection{Verification of Feature Fusion Strategy}
Table \ref{table2} presents the results using three fusion strategies including maxout, meanout, and embedding, where maxout outputs the maximum values of the corresponding positions in the feature maps, and embedding is to use $1\times1$ convolution. It can be seen that all three fusion strategies improved the classification performance. Different from maxout and meanout, when using embedding, the RM is not rotation-invariant and the AP is the lowest. Although the features extracted by maxout are also rotation-invariant, it may lead to the loss of feature continuity. Compared with maxout, meanout can better preserve the features of the samples in different rotation angles. Therefore, we choose meanout as the fusion strategy in our RM network.

\begin{table}[!t]
	\centering
	\caption{\label{table2}The classification results of different fusion strategy}.
	\renewcommand{\arraystretch}{1.2}
	\begin{tabular}{lccc}
		\hline
		Network                & \multicolumn{1}{c}{AP (\%)} & \multicolumn{1}{c}{Ave Sen (\%)} & \multicolumn{1}{c}{Ave Spec (\%)} \\ \hline
		Baseline               & 55.41$\pm$0.95                       & 72.07$\pm$1.57                                        & 96.78$\pm$0.17                                        \\
		Embedding & 61.59$\pm$1.75                       & 74.11$\pm$1.39                                        & 96.78$\pm$0.02                                       \\
	    Maxout    & 62.21$\pm$1.93                       & 72.67$\pm$1.79                                        & 96.89$\pm$0.10                                        \\
		Meanout   & \textbf{64.66$\pm$0.99}                       & \textbf{75.91$\pm$0.81}                                       & \textbf{97.16$\pm$0.04}                                       \\ \hline
	\end{tabular}
\end{table}

\subsubsection{Verification of Rotation Intervals}
 Table \ref{table4} gives the classification results using different rotation intervals $ \theta $, including 90, 60 and 45 degrees. We use bilinear interpolation for rotating. For the interval $ \theta\neq90^\circ $, we pad the boundary region of the feature maps using zero when performing clockwise rotation and then removed the uselessly surrounding region when performing rotating back (realignment), to avoid the information loss caused by rotation.

From Table \ref{table4}, it can be seen that compared with baseline, RM with different rotation intervals all improve the performance of the network. However, the results did not improve significantly as the rotation angle decreased. Theoretically, when the interval becomes smaller, the RM network can extract more information from different angles, which will improve the performance of the network. But in practise, with the increase of rotation angle, interpolation error is introduced during expansion and realignment stages, and the feature extraction ability of CNN is fluctuant with the rotation interval $\theta$ decreasing. In addition, The influence of the interpolation error will also be related to the size of the feature map where RM is located. The feature maps with small size are more easily influenced by the error than the big ones. This is why in Table \ref{table4}, 45-degree intervals do not have a significant advantage over 90-degree. Considering a non-90-degree rotation interval increases the calculation complexity and causes interpolation error, while the classification improvement of a 90-degree is also significant and easy to be implemented, we choose 90-degree as the rotation interval in the later experiments.

\begin{table}[!t]
	\centering
	\caption{\label{table4}The classification results of different rotation intervals}
	\renewcommand{\arraystretch}{1.2}
	\begin{tabular}{lccc}
		\hline
		Network                & \multicolumn{1}{c}{AP (\%)} & \multicolumn{1}{c}{Ave Sen (\%)} & \multicolumn{1}{c}{Ave Spec (\%)} \\ \hline
		Baseline               & 55.41$\pm$0.95                       & 72.07$\pm$1.57                                        & 96.78$\pm$0.17                                        \\
		90 & 64.66$\pm$0.99                       & \textbf{75.97$\pm$0.81}                                        & \textbf{97.16$\pm$0.04}                                       \\
	    60    & 64.47$\pm$0.41                       & 73.85$\pm$0.98                                        & 96.87$\pm$0.12                                        \\
	    45   & \textbf{64.92$\pm$0.76}                       & 73.97$\pm$0.58                                       & 96.84$\pm$0.07                                       \\ \hline
	\end{tabular}
\end{table}

\subsection{Embedding an RM in Different CNNs}
\subsubsection{Embedding an RM in Different Stages of CNNs}
RM is a general operation and can be easily added to different networks. Table \ref{table6} gives the classification results when embedding an strict RM-GAP structure in different CNNs, where $ i_j $ denotes combining an RM from the starting stage $ i^th $ to the last convolution structure $j^th$ before GAP (ResNet34 contains 16 residual blocks, GoogleNet-Inception-V3 contains five types of Inception modules from A to E and EfficientNet-b0 contains 9 stages where stage 2 to 8 are MBConv modules). It can be seen that, after embedding an RM, these networks obtained better performance than baselines, which demonstrates that our RM structure can indeed extract more effective features. Moreover, inserting RM from the middle stage of the network can obtain better results. This is because different starting positions and sizes of feature maps make the RM network get diverse levels of semantic information. In the shallow layers, the extracted features are relatively low-level, while in the deep layers, feature maps have small spatial size and the accurate spatial information is lost. Relatively, the feature maps in middle layers of the network contain appropriate semantic information and their sizes are also moderate. Therefore, the middle layers of CNNs may be better positions to be combined with the RM.

\begin{table}[!t]
	\centering
	\caption{\label{table6}The classification results of embedding a strict RM-GAP structure in different CNNs.}
	\renewcommand{\arraystretch}{1.2}
	\begin{tabular}{lccc}
		\hline
		Network                & \multicolumn{1}{c}{AP (\%)} & \multicolumn{1}{c}{Ave Sen (\%)}& \multicolumn{1}{c}{Ave Spec (\%)} \\ \hline
		\multicolumn{4}{c}{ResNet18}                                                                                                                                                                                           \\ \hline
		Baseline        & 55.41$\pm$0.95   & 72.07$\pm$1.57     & 96.78$\pm$0.17                                                                                   \\
		1\_8             & 60.70$\pm$2.16   & 74.14$\pm$2.43      & 97.00$\pm$0.16                                                                                   \\
		3\_8              & 62.51$\pm$2.17   & 74.34$\pm$2.00       & 96.93$\pm$0.09                                                                                   \\
		5\_8              & \textbf{64.66$\pm$0.18}   & \textbf{75.97$\pm$0.81}      & \textbf{97.16$\pm$0.04}                                                                                   \\
		7\_8            & 58.60$\pm$0.23   & 73.61$\pm$1.73       & 96.87$\pm$0.06                                                                                   \\\hline
		\multicolumn{4}{c}{ResNet34}                                                                                                                                                                                        \\ \hline
		Baseline     & 60.28$\pm$2.33   & 72.61$\pm$1.67     & 96.90$\pm$0.03                                                                                   \\
		1\_16              & 63.19$\pm$2.43   & 69.98$\pm$2.15       & 96.95$\pm$0.08                                                                                   \\
		4\_16              & 63.49$\pm$1.96   & 73.11$\pm$2.05       & 97.07$\pm$0.15                                                                                   \\
		8\_16              & \textbf{67.09$\pm$0.67}   & 74.32$\pm$1.12   & 97.08$\pm$0.02                                                                                  \\
		12\_16              & 63.60$\pm$2.30   & \textbf{75.37$\pm$2.13}       & 97.10$\pm$0.06                                                                                   \\
		14\_16              & 62.09$\pm$0.63   & 71.29$\pm$2.87      & \textbf{97.21$\pm$0.47}                                                                                   \\ \hline
		\multicolumn{4}{c}{GoogleNet-Inception-V3}                                                                                                                                                                          \\ \hline
		Baseline & 59.25$\pm$4.83   & 67.87$\pm$3.96    & 96.30$\pm$0.59                                                                                   \\
		A\_E          & 66.52$\pm$0.72   & 73.52$\pm$1.04   & 96.86$\pm$0.43                                                                                   \\
		B\_E          & 63.34$\pm$0.90   & 68.43$\pm$3.75      & 95.91$\pm$0.57                                                                                   \\
		C\_E          & \textbf{68.86$\pm$2.05}   & \textbf{73.79$\pm$1.69}    & \textbf{97.27$\pm$0.11}                                                                                   \\
		D\_E        & 68.02$\pm$0.55   & 71.12$\pm$0.87  & 97.03$\pm$0.03                                                                                   \\ \hline
		\multicolumn{4}{c}{EfficientNet-b0}                                                                                                                                                                                           \\ \hline
		Baseline        & 67.22$\pm$1.12   & 76.24$\pm$0.80     & 97.51$\pm$0.03                                                                                   \\
		2\_9              & 68.77$\pm$1.32   & 74.24$\pm$2.99      & 97.16$\pm$0.19                                                                                   \\
		4\_9              & 69.09$\pm$0.98   & 75.77$\pm$2.43      & 97.34$\pm$0.11                                                                                   \\
		5\_9              & \textbf{69.67$\pm$0.77}   & \textbf{78.15$\pm$1.02}     & 97.41$\pm$0.08                                                                                   \\
		6\_9            & 68.73$\pm$1.45   & 77.38$\pm$0.79       & 97.52$\pm$0.01                                                                                   \\
		8\_9            & 67.93$\pm$0.74        &76.89$\pm$2.38    & \textbf{97.54$\pm$0.05}                                                                                         \\ \hline
	\end{tabular}
\end{table}

\subsubsection{Comparison Between Relaxed RM-GAP Structure and Strict RM-GAP Structure}
Figure \ref{f2} (b) shows a relaxed RM-GAP structure. Although the structure is not strictly rotation-invariant, it can still improve the anti-rotation ability of the network. In this part, we compare the effects of relaxed RM-GAP structure and strict RM-GAP structure by embedding RM operations into different CNNs, and select the best result of the two methods for comparison. The experiment results in Table \ref{table5} show that embedding the relaxed RM structure into the network also improves the feature extraction ability of the CNNs, but the improvement effect is not as good as the strict one. Because when embedding a relaxed RM-GAP structure, the features extracted by CNNs only have rotation equivariance, which will lead to a low robustness to diverse orientations.

\begin{table}[!t]
	\centering
	\caption{\label{table5}The classification results compared relaxed RM-GAP structure and strict RM-GAP structure in different CNNs.}
	\renewcommand{\arraystretch}{1.2}
	\begin{tabular}{lccc}
		\hline
	Network                & \multicolumn{1}{c}{AP (\%)} & \multicolumn{1}{c}{Ave Sen (\%)}& \multicolumn{1}{c}{Ave Spec (\%)} \\ \hline
		\multicolumn{4}{c}{ResNet18}                                                                                                                                                                                           \\ \hline
		Baseline        & 55.41$\pm$0.95   & 72.07$\pm$1.57     & 96.78$\pm$0.17                                                                                   \\
		Relax             & 58.12$\pm$1.38   & 65.70$\pm$0.88      & 96.31$\pm$0.06                                                                                   \\
		Strict              & \textbf{64.66$\pm$0.99}   & \textbf{75.97$\pm$0.81}       & \textbf{97.16$\pm$0.04}                                                                                   \\\hline
		\multicolumn{4}{c}{ResNet34}                                                                                                                                                                                        \\ \hline
		Baseline     & 59.61$\pm$1.55   & 71.94$\pm$0.99     & 96.90$\pm$0.03                                                                                   \\
		Relax              & 60.18$\pm$0.13   & 73.97$\pm$0.41       & 96.96$\pm$0.04                                                                                   \\
		Strict              & \textbf{67.09$\pm$0.67}   & \textbf{74.32$\pm$1.12}       & \textbf{97.08$\pm$0.02}                                                                                   \\\hline
		\multicolumn{4}{c}{GoogleNet-Inception-V3}                                                                                                                                                                          \\ \hline
		Baseline & 59.25$\pm$4.83   & 67.87$\pm$3.96    & 96.30$\pm$0.59                                                                                   \\
		Relax          & 62.89$\pm$0.03   & 70.29$\pm$0.78   & 96.60$\pm$0.07                                                                                   \\
		Strict          & \textbf{68.86$\pm$2.05}   & \textbf{73.79$\pm$1.69}      & \textbf{97.27$\pm$0.11}                                                                                   \\ \hline
		\multicolumn{4}{c}{EfficientNet-b0}                                                                                                                                                                                           \\ \hline
		Baseline        & 67.22$\pm$1.12   & 76.24$\pm$0.80     & \textbf{97.51$\pm$0.03}                                                                                   \\
		Relax              & 69.14$\pm$0.97   & 74.99$\pm$1.15      & 97.30$\pm$0.05                                                                                   \\
		Strict              & \textbf{69.67$\pm$0.77}   & \textbf{78.15$\pm$1.02}      & 97.41$\pm$0.08                                                                                   \\\hline
	\end{tabular}
\end{table}

\begin{table*}[!t]
\centering
\caption{\label{table8}The classification results of different rotation invariance or equivariance methods(\%).}
\renewcommand{\arraystretch}{1.2}
\resizebox{\textwidth}{!}{
\begin{tabular}{c|c|cccccccc|c}
\hline
Method                                    & Metric      & MEL                     & NV                      & BCC                     & AK                      & BKL                     & DF                      & VASC                    & SCC                     & Average                 \\ \hline
\multirow{4}{*}{Baseline}                 & precision   & 68.88$\pm$2.11          & 94.51$\pm$0.69          & 87.65$\pm$1.68          & 29.07$\pm$2.01          & 61.58$\pm$4.99          & 14.49$\pm$2.51          & 64$\pm$6.93             & 23.12$\pm$0.93          & 55.41$\pm$0.95          \\
                                          & sensitivity & 76.43$\pm$1.65          & 85.47$\pm$1.82          & 77.39$\pm$0.97          & 53.96$\pm$1.20          & 70.42$\pm$2.53          & 53.33$\pm$5.77          & 96.29$\pm$3.21          & 63.24$\pm$2.20          & 72.07$\pm$1.57          \\
                                          & specificity & 93.37$\pm$0.38          & 93.62$\pm$0.57          & 98.09$\pm$0.26          & 97.54$\pm$0.06          & 95.63$\pm$0.52          & 99.22$\pm$0.02          & 99.64$\pm$0.07          & 98.10$\pm$0.02          & 96.78$\pm$0.17          \\
                                          & kappa       & -                       & -                       & -                       & -                       & -                       & -                       & -                       & -                       & 70.39$\pm$1.07          \\ \hline
\multirow{4}{*}{RA}                       & precision   & 64.82$\pm$2.55          & 95.28$\pm$0.75          & \textbf{89.26$\pm$2.12} & 29.46$\pm$3.36          & 59.03$\pm$0.79          & 18.84$\pm$6.64          & 74.67$\pm$2.31          & 27.96$\pm$0.93          & 57.42$\pm$0.56          \\
                                          & sensitivity & 77.98$\pm$2.55          & 84.18$\pm$0.77          & 76.13$\pm$1.01          & 58.85$\pm$3.52          & 73.57$\pm$3.59          & 75.79$\pm$9.54          & 93.41$\pm$2.54          & 65.97$\pm$8.42          & 74.42$\pm$1.77          \\
                                          & specificity & 92.62$\pm$0.44          & \textbf{94.35$\pm$0.80} & \textbf{98.34$\pm$0.32} & 97.36$\pm$0.11          & 95.37$\pm$0.06          & 99.26$\pm$0.06          & 99.75$\pm$0.02          & 98.21$\pm$0.02          & 96.85$\pm$0.10          \\
                                          & kappa       & -                       & -                       & -                       & -                       & -                       & -                       & -                       & -                       & 70.03$\pm$0.51          \\ \hline
\multirow{4}{*}{TI-pooling}               & precision   & 68.21$\pm$4.66          & 93.50$\pm$0.70          & 87.95$\pm$2.11          & 31.78$\pm$7.01          & 62.09$\pm$1.76          & 28.99$\pm$5.02          & 74.67$\pm$2.31          & 27.42$\pm$2.79          & 59.33$\pm$1.43          \\
                                          & sensitivity & 75.42$\pm$0.68          & 85.89$\pm$1.21          & 75.83$\pm$2.39          & 51.97$\pm$3.51          & 73.67$\pm$4.65          & 65.51$\pm$8.25          & 90.53$\pm$4.19          & 61.86$\pm$3.71          & 72.58$\pm$0.75          \\
                                          & specificity & 93.23$\pm$0.89          & 92.59$\pm$0.62          & 98.14$\pm$0.32          & 97.63$\pm$0.23          & 95.69$\pm$0.16          & 99.31$\pm$0.10          & 99.75$\pm$0.02          & 98.20$\pm$0.07          & 96.82$\pm$0.08          \\
                                          & kappa       & -                       & -                       & -                       & -                       & -                       & -                       & -                       & -                       & 70.41$\pm$1.30          \\ \hline
\multirow{4}{*}{GCNN}                     & precision   & 61.06$\pm$2.61          & 95.26$\pm$0.71          & 87.15$\pm$1.14          & 22.48$\pm$4.08          & 56.36$\pm$1.81          & 10.15$\pm$2.51          & \textbf{82.67$\pm$2.31} & 16.13$\pm$1.61          & 53.19$\pm$0.53          \\
                                          & sensitivity & 78.26$\pm$2.78          & 82.46$\pm$1.06          & 73.33$\pm$0.92          & 48.55$\pm$9.46          & 69.75$\pm$1.23          & 69.45$\pm$4.81          & 96.97$\pm$5.25          & 62.05$\pm$10.96         & 72.60$\pm$1.30          \\
                                          & specificity & 91.91$\pm$0.44          & 94.16$\pm$0.70          & 98.00$\pm$0.17          & 97.32$\pm$0.13          & 95.07$\pm$0.20          & 99.18$\pm$0.02          & 99.83$\pm$0.02          & 97.93$\pm$0.04          & 96.58$\pm$0.10          \\
                                          & kappa       & -                       & -                       & -                       & -                       & -                       & -                       & -                       & -                       & 66.23$\pm$1.10          \\ \hline
\multirow{4}{*}{ORN}                      & precision   & 59.81$\pm$1.79          & 95.15$\pm$0.31          & 86.34$\pm$1.52          & 22.09$\pm$1.17          & 51.40$\pm$1.54          & 13.04$\pm$4.35          & 62.67$\pm$2.31          & 23.12$\pm$1.86          & 51.70$\pm$1.21          \\
                                          & sensitivity & 75.61$\pm$2.91          & 81.81$\pm$0.39          & 71.82$\pm$1.50          & 48.85$\pm$2.24          & 69.43$\pm$0.81          & 69.45$\pm$4.81          & 96.08$\pm$6.79          & 68.51$\pm$8.35          & 72.45$\pm$2.54          \\
                                          & specificity & 91.63$\pm$0.32          & 93.96$\pm$0.32          & 97.87$\pm$0.23          & 97.31$\pm$0.04          & 94.55$\pm$0.15          & 99.21$\pm$0.04          & 99.63$\pm$0.02          & 98.10$\pm$0.05          & 96.49$\pm$0.13          \\
                                          & kappa       & -                       & -                       & -                       & -                       & -                       & -                       & -                       & -                       & 65.27$\pm$1.03          \\ \hline
\multirow{4}{*}{STN}                      & precision   & 65.16$\pm$2.29          & 95.16$\pm$0.38          & 86.84$\pm$2.22          & 34.49$\pm$4.08          & 55.62$\pm$5.36          & 14.49$\pm$2.51          & 72.67$\pm$3.06          & 23.04$\pm$3.33          & 55.93$\pm$1.31          \\
                                          & sensitivity & 75.58$\pm$1.15          & 81.64$\pm$3.33          & 75.13$\pm$2.48          & 53.91$\pm$3.38          & 74.22$\pm$5.17          & 86.67$\pm$23.09         & 94.73$\pm$0.28          & 66.20$\pm$10.10         & 74.13$\pm$1.77          \\
                                          & specificity & 92.46$\pm$0.48          & 94.01$\pm$0.17          & 97.82$\pm$0.36          & 97.72$\pm$0.14          & 94.67$\pm$0.67          & 99.22$\pm$0.02          & 99.72$\pm$0.04          & 98.07$\pm$0.08          & 96.71$\pm$0.16          \\
                                          & kappa       & -                       & -                       & -                       & -                       & -                       & -                       & -                       & -                       & 67.65$\pm$2.27          \\ \hline
\multirow{4}{*}{CyCNN}                    & precision   & \textbf{72.86$\pm$1.05} & 93.53$\pm$0.65          & 88.86$\pm$0.91          & 29.46$\pm$1.78          & 64.12$\pm$0.66          & 13.71$\pm$1.15          & 78.67$\pm$6.11          & 31.05$\pm$3.58          & 59.03$\pm$0.95          \\
                                          & sensitivity & 79.28$\pm$1.85          & 87.38$\pm$1.10          & 73.15$\pm$0.62          & 47.03$\pm$4.10          & \textbf{75.35$\pm$1.20} & 60.67$\pm$1.15          & 94.07$\pm$5.14          & 67.25$\pm$2.74          & 73.02$\pm$1.30          \\
                                          & specificity & \textbf{94.19$\pm$0.18} & 92.77$\pm$0.56          & 98.26$\pm$0.14          & 97.55$\pm$0.06          & 95.93$\pm$0.07          & 99.22$\pm$0.02          & 99.79$\pm$0.06          & 98.29$\pm$0.08          & 97.00$\pm$0.02          \\
                                          & kappa       & -                       & -                       & -                       & -                       & -                       & -                       & -                       & -                       & 72.21$\pm$0.32          \\ \hline
\multirow{4}{*}{E$^4$-Net} & precision   & 65.42$\pm$1.59          & 94.94$\pm$1.84          & 85.87$\pm$0.90          & 29.24$\pm$2.29          & 53,50$\pm$3.60          & 18.72$\pm$1.53          & 69.23$\pm$7.37          & 20.15$\pm$2.11          & 54.69$\pm$1.70          \\
                                          & sensitivity & 75.46$\pm$0.58          & 82.51$\pm$0.51          & 71.55$\pm$0.96          & 52.13$\pm$1.77          & 68.52$\pm$1.76          & \textbf{91.67$\pm$8.33} & 94.41$\pm$0.57          & 65.26$\pm$7.39          & 75.15$\pm$0.76          \\
                                          & specificity & 92.43$\pm$0.18          & 92.67$\pm$0.33          & 97.60$\pm$0.10          & 97.50$\pm$0.07          & 94.62$\pm$0.21          & 99.27$\pm$0.03          & 99.68$\pm$0.07          & 98.00$\pm$0.08          & 96.47$\pm$0.03          \\
                                          & kappa       & -                       & -                       & -                       & -                       & -                       & -                       & -                       & -                       & 69.74$\pm$3.13          \\ \hline
\multirow{4}{*}{RiDe}                     & precision   & 62.98$\pm$1.57          & \textbf{95.72$\pm$0.81} & 82.23$\pm$0.79          & 31.78$\pm$2.42          & 54.45$\pm$2.48          & 15.94$\pm$2.51          & 56.00$\pm$6.93          & 26.88$\pm$0.93          & 53.17$\pm$0.77          \\
                                          & sensitivity & 78.18$\pm$1.64          & 82.41$\pm$0.99          & 73.99$\pm$1.18          & 52.00$\pm$3.46          & 68.59$\pm$1.81          & 65.71$\pm$12.45         & 79.09$\pm$4.17          & 61.85$\pm$3.39          & 70.23$\pm$1.05          \\
                                          & specificity & 92.27$\pm$0.27          & 93.91$\pm$0.35          & 97.27$\pm$0.12          & 97.63$\pm$0.08          & 94.86$\pm$0.27          & 99.24$\pm$0.02          & 99.56$\pm$0.07          & 98.19$\pm$0.02          & 96.61$\pm$0.04          \\
                                          & kappa       & -                       & -                       & -                       & -                       & -                       & -                       & -                       & -                       & 66.76$\pm$0.78          \\ \hline
\multirow{4}{*}{RIR}                      & precision   & 69.62$\pm$1.47          & 94.66$\pm$1.37          & 86.54$\pm$1.55          & 39.92$\pm$2.92          & 58.91$\pm$6.46          & 20.29$\pm$5.02          & 70.67$\pm$6.11          & 27.42$\pm$7.03          & 58.50$\pm$1.20          \\
                                          & sensitivity & 77.71$\pm$1.51          & 84.63$\pm$1.28          & 78.52$\pm$0.88          & 53.52$\pm$4.20          & 75.31$\pm$3.95          & 58.09$\pm$1.65          & \textbf{98.33$\pm$2.89} & \textbf{76.21$\pm$1.42} & 75.29$\pm$1.30          \\
                                          & specificity & 93.53$\pm$0.30          & 93.73$\pm$1.36          & 97.94$\pm$0.23          & 97.90$\pm$0.09          & 95.37$\pm$3.95          & 99.28$\pm$0.05          & 99.71$\pm$0.06          & 76.21$\pm$1.42          & 96.99$\pm$0.19          \\
                                          & kappa       & -                       & -                       & -                       & -                       & -                       & -                       & -                       & -                       & 70.85$\pm$0.88          \\ \hline
\multirow{4}{*}{RM}                       & precision   & 71.61$\pm$2.68          & 93.81$\pm$0.24          & 87.65$\pm$1.51          & \textbf{40.70$\pm$3.07} & \textbf{69.21$\pm$5.56} & \textbf{34.78$\pm$4.35} & 78.67$\pm$8.33          & \textbf{40.86$\pm$8.12} & \textbf{64.66$\pm$0.99} \\
                                          & sensitivity & \textbf{79.51$\pm$3.42} & \textbf{87.58$\pm$1.01} & \textbf{79.25$\pm$1.24} & \textbf{59.72$\pm$4.16} & 74.45$\pm$3.16          & 65.28$\pm$4.32          & 97.22$\pm$4.81          & 64.72$\pm$3.60          & \textbf{75.97$\pm$0.81} \\
                                          & specificity & 93.96$\pm$0.47          & 93.08$\pm$0.17          & 98.10$\pm$0.23          & \textbf{97.94$\pm$0.10} & \textbf{96.47$\pm$0.59} & \textbf{99.40$\pm$0.04} & \textbf{99.79$\pm$0.08} & \textbf{98.53$\pm$0.20} & \textbf{97.16$\pm$0.04} \\
                                          & kappa       & -                       & -                       & -                       & -                       & -                       & -                       & -                       & -                       & \textbf{73.91$\pm$0.57} \\ \hline
\end{tabular}}
\end{table*}

\subsection{Comparison with Other Rotation Invariance or Equivariance Methods}
Taking the ResNet18 as baseline, we compare our RM network with other state-of-the-art rotation invariance or equivariance methods including rotation data augment (RA), TI-pooling \cite{laptev2016ti}, GCNN\cite{cohen2016group},ORN\cite{zhou2017oriented}, STN\cite{jaderberg2015spatial}, CyCNN\cite{kim2020cycnn}, E$^4$-Net\cite{he2021efficient}, RiDe\cite{kang2021rotation}, and RIR\cite{qi2021rotation} in dermoscopy image classification task. For a fair comparison, we retrain these compared methods on ISIC 2019 dataset and all the rotation intervals of these methods are set to 90-degree. In addition, similar to our RM, GCNN, ORN, STN and E$^4$-Net models can also be easily inserted into different parts of the networks, thus we select the best classification results here. Table \ref{table8} gives the classification results on the testing dataset.

\begin{figure*}[!t]
	\centering
	\includegraphics[width=1\textwidth]{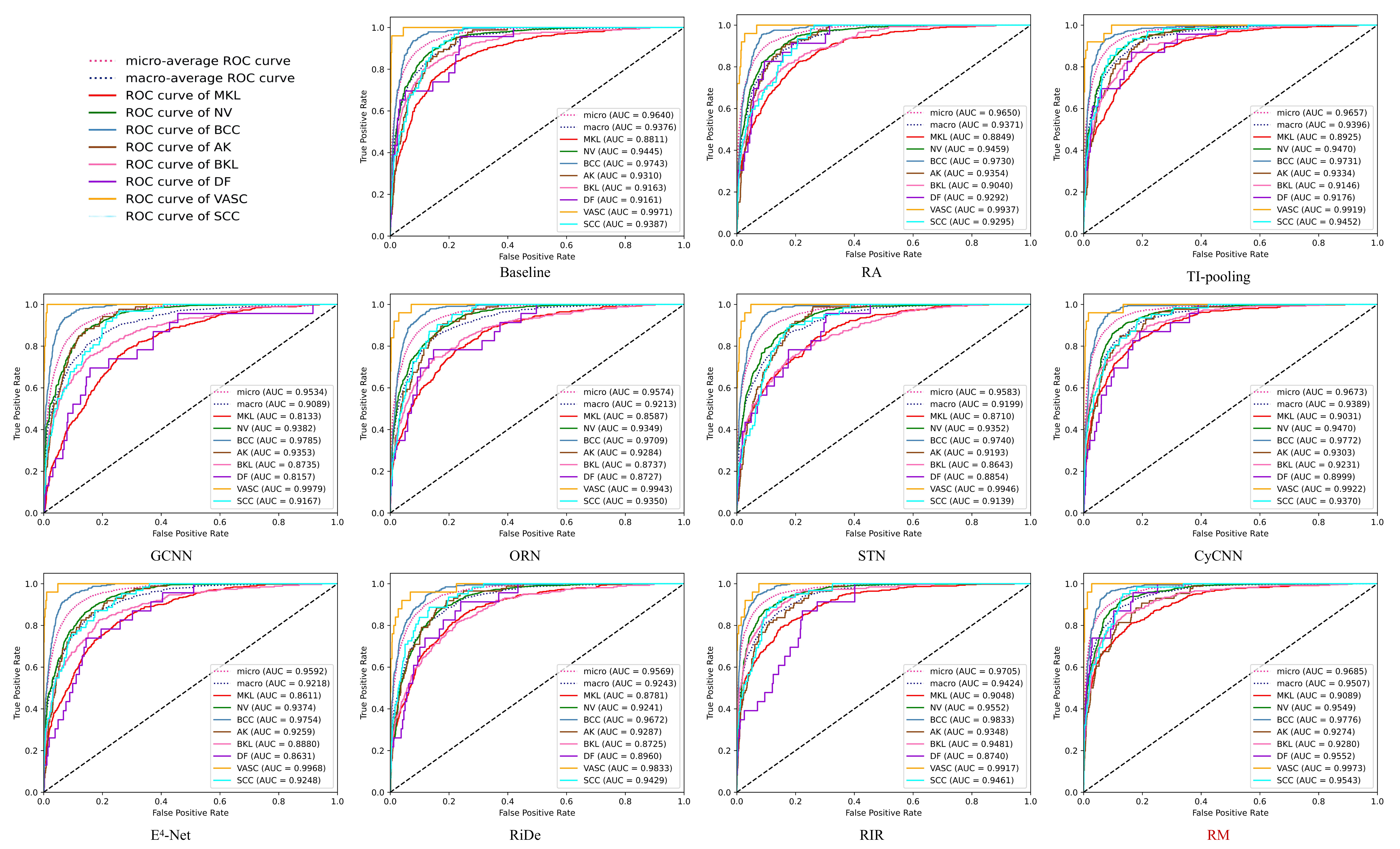}
	\caption{ROC curves of different anti-rotation methods: The figure shows the ROC curves and the AUC of the overall model and 8 types of skin disease classification results.}
	\label{f5}
\end{figure*}

From the results, we can find that DA-based methods, including RA, TI-pooling, CyCNN, and RIR, can improve the performance of the network except for RiDe method. This kind of method is trained with a dataset enriched by rotation, making each class contain possible orientations. However, DA-based methods come at the price of a higher probability of overfitting and a significant increase of computational complexity. For example, when the rotation angle of inputs is 90 degrees, the training time of TI-Pooling is theoretically increased by four times. We can also see that methods based on rotated filters, including GCNN, ORN, and E$^4$-Net, get lower metrics. This could be attributed to extra parameters when these methods are inserted into a layer of CNNs, resulting in difficulty in using transfer learning of the changed layers. Besides, Anti-transformation method STN get a better performance than baseline, but it is not mainly designed for rotation and there is no strict rotation invariance, leading to a limited improvement. According to experimental results, our method achieves superiority on most categories in terms of the precision, sensitivity, specificity and kappa metrics.

To further display the superiority over other methods, we also draw the receiver operating characteristic (ROC) curve and report the area under the curve (AUC) of the overall model and 8 types of skin disease, shown in Figure \ref{f5}. It can be seen that the ROC curves of our method are all closer to the upper left corner of the distance. Relative to the compared methods, the ROC curves of our RM network are more compact, which proves that RM can effectively improve the rotation-invariant feature extraction ability of the network. For the problem of the imbalanced dataset, the classification performance of a small number of categories can be improved, achieving better performance on the classification task. The AP of our method is 5.63\% higher than the second place of compared methods. At the same time, the Kappa consistency has also been greatly improved, close to a high degree (Kappa 0.7391, close to 0.75). Therefore, with significant improvement, our method outperformed the compared methods.

\subsection{Comparison with Other Dermoscopy Image Classification Methods}

From the above experiments, we have used the classic CNNs and compared them with the network after inserting RM. In this part of the experiments, we compare some recent methods designed for skin lesion classification task. We chose the classification method based on the basic network, ARLNets\cite{zhang2019attention} and RegNets\cite{yao2021single,radosavovic2020designing}. ARLNets are designed based on ResNets for dermoscopy classification. The residual attention module improves the network's ability to represent skin lesions. Recently, Yao \etal\cite{yao2021single} used the RegNets\cite{radosavovic2020designing} to extract features for the first time in the dermoscopic image classification task, and achieved better performance than other classic CNNs, and designed a multi-weighted new loss (MWNL) method for imbalanced small dermoscopy datasets.

To ensure the fairness of comparison, we retrain the networks of compared methods on ISIC2019 dataset. In addition, we use the same baseline for different methods and embed the RM network into them. In the comparison of the first three groups in Table \ref{table9}, we use the same setup as the above experiments. It can be seen that our method performs better. Then, in the last group, RegNetY800M\_MWNL means using the MWNL loss function for training, and the training details are the same as that in \cite{yao2021single}. We can also find that applying our RM to this method can also obtain better results, which indicate that RM has great universality and generality.

\begin{table*}[!t]
\centering
\caption{\label{table9}The performance of other dermoscopy image classification methods.}
\renewcommand{\arraystretch}{1.2}
	\setlength\tabcolsep{15pt}
    \begin{tabular}{lcccc}
    \hline
	Network                & \multicolumn{1}{c}{AP (\%)} & \multicolumn{1}{c}{Ave Sen (\%)}& \multicolumn{1}{c}{Ave Spec (\%)} & \multicolumn{1}{c}{Kappa (\%)}\\ \hline
    ARLNet18       &   59.53$\pm$0.36     & 73.31$\pm$0.64  &     96.90$\pm$0.02  &   71.10$\pm$0.19    \\
    ResNet18\_RM        &\textbf{64.66$\pm$0.99}        &\textbf{75.97$\pm$0.81} &\textbf{97.16$\pm$0.04}   &\textbf{74.12$\pm$0.83}       \\ \hline
    ARLNet34 & 60.21$\pm$1.06       &    73.59$\pm$1.17 & 96.64$\pm$0.10 &  69.53$\pm$0.64     \\
    ResNet34\_RM       &\textbf{67.09$\pm$0.67}  &\textbf{74.32$\pm$1.12}  &\textbf{97.08$\pm$0.02}  &\textbf{73.39$\pm$0.35}       \\ \hline
    RegNetY800M & 64.32$\pm$1.21       &     76.84$\pm$0.25   &  97.23$\pm$0.02   & 73.60$\pm$0.06       \\
    RegNetY800M\_RM & \textbf{69.53$\pm$1.31}  & \textbf{77.73$\pm$0.57}  & \textbf{97.26$\pm$0.16}  & \textbf{75.24$\pm$0.91} \\ \hline
    RegNetY800M\_MWNL& 77.82$\pm$0.78       & 77.62$\pm$0.93 &  97.35$\pm$0.04   & 77.10$\pm$0.32       \\
    RegNetY800M\_MWNL\_RM & \textbf{79.85$\pm$1.55}  & \textbf{81.62$\pm$2.39}  & \textbf{97.59$\pm$0.14}  & \textbf{79.33$\pm$0.76} \\ \hline
    \end{tabular}
\end{table*}

\subsection{Using RM in Dermoscopy Image Retrieval Task}
Dermoscopy image retrieval technology is an effective method to assist doctors in diagnosis. Doctors can get some reference information for diagnosis by retrieving similar dermoscopy images and their diagnosis reports in the database. The proposed RM can also be applied to the retrieval task of dermoscopy images.

\subsubsection{Verification of RM in Retrieval Task}
We also use the same CNNs including ResNet18, ResNet34, GoogleNet-Inception-V3, and EfficientNet-b0 as baselines. The embedding position of the RM network is the same as the best performing network in Table \ref{table5}. The training set of ISIC 2019 is used to build the database, and the images in the testing set are the query images. Table \ref{table7} displays the image retrieval results evaluated by both mAP and mRR.  Consistently with the classification results, RM exhibits superior performance when compared to the original CNNs, which indicates that the network with the RM inserted can extract deep hash codes with rotation invariance.

\begin{table}[!t]
	\centering
	\caption{\label{table7}The retrieval performance of RM networks in different CNNs.}
	\renewcommand{\arraystretch}{1.2}
	\begin{tabular}{lcc}
		\hline
		Network                & mAP@10(\%) & mRR@10(\%) \\ \hline
		ResNet18               & 54.11$\pm$0.29                         & 63.56$\pm$1.60                        \\
		ResNet18\_RM      & \textbf{60.20$\pm$2.74}                         &\textbf{67.82$\pm$2.48}                         \\
		ResNet34               & 57.18$\pm$3.80                         & 66.46$\pm$2.21                         \\
		ResNet34\_RM     & \textbf{59.67$\pm$4.77}                         & \textbf{68.16$\pm$2.68}                         \\
		Inception-V3          & 59.51$\pm$0.92                         & 66.74$\pm$0.52                         \\
		Inception-V3\_RM & \textbf{63.36$\pm$3.46}                         & \textbf{70.50$\pm$3.24}                         \\ 
		EfficientNet-b0          & 68.47$\pm$0.33                         & 76.88$\pm$0.66                         \\
		EfficientNet-b0\_RM & \textbf{69.84$\pm$0.47}                         & \textbf{77.16$\pm$0.87}                         \\
		\hline
	\end{tabular}
\end{table}

\subsubsection{Comparison with Other Retrieval Methods}
We compare other retrieval methods for natural images and dermoscopy images, including natural image retrieval methods DCH \cite{Cao_2018_CVPR}, DPN\cite{fan2020deep} and CSQ \cite{yuan2020central}, and dermoscopy image retrieval method CRI \cite{zhang2021dermoscopic}. The methods above all improve the loss function to make the intra-class distribution of the learned depth hash code more compact. Among them, CRI method also considers that the type of skin lesions does not change due to the imaging angle, and proposes an anti-rotation loss term based on the Cauchy distribution probability function. This method obtains certain rotation invariance by rotating the input image at equal intervals and learning the output differences of samples at different angles. We use ResNet18 as the baseline and the method of constructing a deep hashing network is the same as the previous experiment.  

\begin{table}[!t]
	\centering
	\caption{\label{other_retrieval}The performance of other image retrieval methods.}
	\renewcommand{\arraystretch}{1.2}
	\setlength\tabcolsep{15pt}
	\begin{tabular}{lcc}
		\hline
		Network                & mAP@10(\%) & mRR@10(\%) \\ \hline
		Baseline               & 53.75$\pm$0.40                         & 62.85$\pm$0.73                         \\
		DCH      & 53.11$\pm$1.80                         & 62.95$\pm$1.03                         \\
		DPN               & 56.05$\pm$0.76                         & 64.94$\pm$1.01                         \\
		CSQ     & 55.68$\pm$0.11                         & 65.15$\pm$1.04                        \\
		CRI          & 58.67$\pm$1.49                         & 65.15$\pm$0.32                         \\
		RM & \textbf{60.20$\pm$2.74}                         & \textbf{67.82$\pm$2.48}                         \\ \hline
	\end{tabular}
\end{table}
Table \ref{other_retrieval} presents the image retrieval results. It can be observed that RM and CRI perform better than other compared algorithms, which means that the rotation invariance methods can effectively solve the rotation problem caused by the unfixed shooting angles of images. While DCH, DPN, and CSQ ignore the characteristics of dermoscopy images. Although they have shown the advantages of natural image dataset, the generality of dermoscopy image retrieval task is still limited. Besides, compared to the second place method CRI, mAP@10 and mRR@10 of our method are improved by 1.53\% and 2.67\% respectively. Although CRI is also based on improving the anti-rotation ability of networks, the need for rotating input images leads to a high computational cost. While our RM network is used for partial convolution structure, compared with the CRI method using the entire siamese network, the computational complexity is lower under the same rotation intervals. This suggests that RM can learn high-quality deep hash codes based on both fewer calculation efforts and powerful CNN architectures. 

\section{Conclusion}
CNNs have been widely used in the assisted diagnosis of dermoscopy images and achieved significant results. However, The characteristic that the lesion target has no principal direction in the imaging process of dermoscopy images leads to a large number of target rotations in the dataset, while the regular CNNs are not invariant to the rotations. In this paper, a general anti-rotation model RM is proposed for dermoscopy images, which can be used in classification and retrieval tasks. 

Through theoretical derivation, the proposed RM can extract rotation-invariant features when being followed by a GAP operation. Our method has a great generality and can be flexibly embedded in any part of a standard CNN, which does not change the network structure or increase the parameters. Extensive experiments on ISIC 2019 demonstrate the effectiveness of our method. Taking widely used CNNs including ResNet18, ResNet34, GoogleNet-Inception-V3, and EfficientNet-b0 as baseline networks respectively, qualitative and quantitative experiments were carried out on classification and retrieval tasks. Through experiments, the influence factors on the RM, such as feature fusion strategy, embedding position, rotation interval, etc., were systematically discussed, which provides a reference to the readers for better using our RM. Compared with other rotation invariance or equivariance methods, our method can more effectively improve the performance and also outperforms the compared algorithms in the classification and retrieval tasks. Our study shows the potential of rotation-invarient features in the field of dermoscopy images.

In addition, we also try to use RM in the skin lesion segmentation task. However, the experimental results show that the improvement of the network is not obvious. Because the network usually does not use GAP layer and rotation equivariance is more important than invariance in segmentation task, we can only embed a relaxed RM structure into CNNs, thus this task is not adopted. The potential of RM in the segmentation task needs to be further explored.

\section{Acknowledgements}
This work was supported by the National Natural Science Foundation of China (Nos. 61871011, 82173449, 62071011 and 61971443).	




\bibliographystyle{elsarticle-num}
\bibliography{manuscript.bib}





\end{document}